\documentclass{article}

\usepackage[english]{babel}

\usepackage[letterpaper,top=2cm,bottom=2cm,left=3cm,right=3cm,marginparwidth=1.75cm]{geometry}
\usepackage{tabularray}
\usepackage{multirow}
\usepackage{xcolor,colortbl}
\usepackage{booktabs}
\usepackage{amsmath}
\usepackage{graphicx}
\usepackage[colorlinks=true, allcolors=blue]{hyperref}

\title{Analyzing Quantization in TVM}
\author{Mingfei Guo(mfguo@stanford.edu)}

\begin{document}
\maketitle

% \begin{abstract}
% Your abstract.
% \end{abstract}
\setcounter{secnumdepth}{4}
\section{Introduction}
\subsection{Background}
\subsubsection{Quantization}
Quantization is a technique commonly used in deep learning frameworks to reduce the precision of neural network weights and activations from 32-bit floating-point numbers to 8-bit integers. The goal of quantization is to reduce computational requirements and decrease memory usage while maintaining acceptable model accuracy. 

\subsubsection{TVM (Tensor Virtual Machine)}
TVM is an open-source compiler stack inspired by Halide that optimizes and deploys deep learning models on various hardware platforms. It aims to enable machine learning engineers to optimize and run computations efficiently on any hardware backend.

\bigskip
\noindent TVM comprises two optimization layers. The first layer focuses on computation graph optimization, addressing high-level dataflow rewriting. The second layer, the tensor optimization layer, introduces new schedule primitives to optimize memory reuse across threads, leverage tensorized compute intrinsics, and improve latency hiding techniques.
For example, optimization schedules involve loop reordering, axis splitting, cache read/write definitions, and more. These schedules enable developers to fine-tune computation execution, harness hardware-specific optimizations, and eliminate the need for manual design.

\subsection{The Problem}

There has been many papers in academic literature on quantizing weight tensors in deep learning models to reduce inference latency and memory footprint. TVM also has the ability to quantize weights and support low-bit computations.
Although quantization is typically expected to improve inference time, in TVM, the performance of 8-bit quantization does not meet the expectations.

\bigskip
\noindent Typically, when applying 8-bit quantization to a deep learning model, it is usually expected to achieve around 50\% of the full-precision inference time. However, in this particular case, not only does the quantized version fail to achieve the desired performance boost, but it actually performs worse, resulting in an inference time that is about 2 times as slow as the non-quantized version.

\subsection{Objective}
In this project, we thoroughly investigate the reasons behind the underperformance and assess the compatibility and optimization opportunities of 8-bit quantization in TVM.
We discuss the optimization of two different types of tasks: computation-bound and memory-bound, and provide a detailed comparison of various optimization techniques in TVM.

\bigskip
\noindent Through the identification of performance issues, we have successfully improved quantization by addressing a bug in graph building. Furthermore, we analyze multiple optimization strategies to achieve the optimal quantization result.
The best experiment achieves 163.88\%
improvement compared with the TVM compiled baseline in inference time for the compute-bound task and 
194.98\% for the memory-bound task.

\section{Methodology}
\subsection{Tasks Definition}
\subsubsection{Computation-Bound Tasks}

Computation-bound tasks refer to situations where the primary limiting factor is the computational resources required to perform the necessary calculations, such as matrix multiplications and convolutions.

\bigskip
\noindent In general, machine learning tasks tend to be more computation-bound due to the significant computational resources required to perform inference or training. We consider batch size 1 tasks as computation-bound tasks because memory is not utilized to a great extent in this scenario.

\bigskip
\noindent We expect to benefit from quantization in computation-bound tasks because INT8 computations involve lower-precision arithmetic compared to FP32 and require fewer resources and less computational complexity.
The hardware generally has specialized arithmetic units designed for floating point math and fixed point math. These units can execute INT8 instructions faster than FP32 instructions.

\subsubsection{Memory-Bound Tasks Definition}

Memory-bound tasks refer to scenarios where the primary limiting factor is the memory bandwidth or the amount of data that can be efficiently moved between memory and the processor. Large input data can cause frequent data transfers between memory and the processor, which can become a performance bottleneck.

\bigskip
\noindent Larger batch sizes tend to be more memory-bound because larger batches require more memory to store the input data, intermediate computations, and gradients. With batch sizes like 8, 64, or 256, there is a higher demand for memory to hold the larger number of samples being processed simultaneously. This can lead to increased memory bandwidth usage and potential memory-related performance limitations.

\bigskip
\noindent By quantizing from 32-bit floating-point to 8-bit integers, the memory required to store model parameters is reduced by a factor of 4. This reduction in precision allows for more compact storage of numerical values, resulting in substantial memory savings. The memory bandwidth required to transfer the data between memory and processing units (e.g., CPU, GPU) is also reduced, which contributes to improved memory utilization.

\subsection{Experimental Setup}
We explore ResNet18 inference compiled by TVM, using different precisions and model layouts.
The system is an aarch64 architecture with an 8-core arm Cortex-A72 CPU, running Ubuntu 22.04.2 LTS with 15.6GB of memory.
For each experiment, we average the performance over 110 epochs with the first 10 epochs used for warm-up. Each epoch consists of inference on a batch of validation data.

\section{Results and Analysis}

\subsection{Bug Fix}

\begin{table}[ht]
       \centering
        \caption{\textbf{ResNet-18 inference result.} PyTorch and TVM represent the original model and the TVM-compiled fp32 model. TVM-Quant refers to the original TVM-compiled int8 quantized model, while TVM-Quant-SG represents our fixed version of the TVM-compiled int8 quantized model. \\}
        \label{tab:sg}
        \begin{tabular}{@{}cccccc@{}}
            \toprule
            \textbf{\begin{tabular}[c]{@{}c@{}}Framework\end{tabular}}     & \textbf{Layout}   & \textbf{\begin{tabular}[c]{@{}c@{}}Schedule\end{tabular}}    & \textbf{Precision} & \textbf{Time (ms)}  & \textbf{\begin{tabular}[c]{@{}c@{}}Improvement\end{tabular}}  \\
            \midrule
             PyTorch                        & NCHW     & -  & fp32 & 69.26  & -  \\
             TVM        & NCHW   & nchw\_spatial\_pack  & fp32 & 13.29  &   100\%  \\
             TVM-Quant        & NCHW   & nchw\_spatial\_pack  & int8 & 29.19 &  45.52\%  \\
             \rowcolor{lightgray} TVM-Quant-Graph        & NCHW   & nchw\_spatial\_pack  & int8 & 8.27 & 160.70\%   \\

             % \midrule
           \bottomrule
        \end{tabular}
        \end{table}

First of all, we fixed the bug that was causing simple quantization to be much slower than the full precision version. Theoretically, at least, quantization should not cause any harm. After fixing the layout and schedule, we found that the quantization performance was still much slower than the pure fp32 TVM compiled version. Therefore, we suspected that the problem existed at the graph level optimization.

\bigskip
\noindent We discovered that TVM provides two types of executors: Graph Executor and VM (Virtual Machine) Executor. The Graph Executor is designed for efficient execution of pre-optimized computation graphs. It takes a static model graph, where every operation is pre-defined, and optimizes it through various graph-level optimizations for the target hardware. However, the VM Executor is a lower-level executor that allows dynamic operations, enabling runtime code generation, and is suitable for dynamic networks such as RNN.

\bigskip
\noindent By default, TVM's quantization code sets the executor to VM, allowing the model to be potentially partitioned into three modules: a prefix function (which converts inputs into the quantized data space), a middle function (which contains the core quantized network), and a suffix function (which dequantizes the output). However, in this experiment, we don't need model partitioning or dynamic allocation during runtime. Therefore, we resolved this issue by using a static graph relay executor. Results shown in TVM-Quant-SG in Table \ref{tab:sg}, indicate that after resetting the graph executor, we achieved a 160.70\% improvement over the original TVM compiled fp32 model. We believe this improvement mainly stems from executing low bit operations, which is quite reasonable. Since we only perform inference on one image per batch in this case, memory shouldn't be a problem.

\subsection{Performance Optimization and Analysis}

\subsubsection{Computation-Bound Tasks}
After fixing the issue, the quantization results seem reasonable to us. Therefore, we attempted to explore tensor-level optimizations using the fixed TVM-Quant-Graph framework to analyze the results. Our focus was primarily on the conv2d strategy as it represents the most computationally intensive task in our model.

\bigskip
\noindent Ideally, we would assume that all optimizations are orthogonal, meaning that applying a new optimization should yield significant improvements. However, this assumption does not hold true in practice because TVM provides predefined schedules that are highly optimized for almost every setting. Consequently, when setting int8 quantization, we might invoke a different strategy, and it would be unfair to expect a direct comparison of its performance with the fp32 version. The results are in Table \ref{tab:schedule}.

\begin{table}[ht]
       \centering
        \caption{\textbf{ResNet-18 inference with batch size 1 under the fixed framework TVM-Quant-Graph.} The optimizations are not orthogonal, as different settings would map to different schedules, and these schedules are optimized to varying degrees. The improvement for computation-bound tasks depends on the quality of the setup and the schedule.  \\}
        \label{tab:schedule}
        \begin{tabular}{@{}ccccc@{}}
            \toprule
            \textbf{\begin{tabular}[c]{@{}c@{}}Layout\end{tabular}}     & \textbf{Schedule}   & \textbf{\begin{tabular}[c]{@{}c@{}}Precision\end{tabular}}    & \textbf{Time (ms)} & \textbf{Ideal Speedup}   \\
            \midrule
            \multirow{3}{*}{NCHW}                                & spatial\_pack  & fp32 & 13.29  &   16x \\
            & spatial\_pack & int8 & 8.27 & 16x \\

            & simd & int8 & 11.36 & 16x \\

            \midrule
            \multirow{2}{*}{NHWC}      & spatial\_pack & fp32  & 35.15  & 4x   \\

            & quantized\_interleaved   & int8 & 12.09  & 16x   \\

           \bottomrule
        \end{tabular}
        \end{table}

\begin{figure}[htp]
    \centering
    \includegraphics[width=12.5cm]{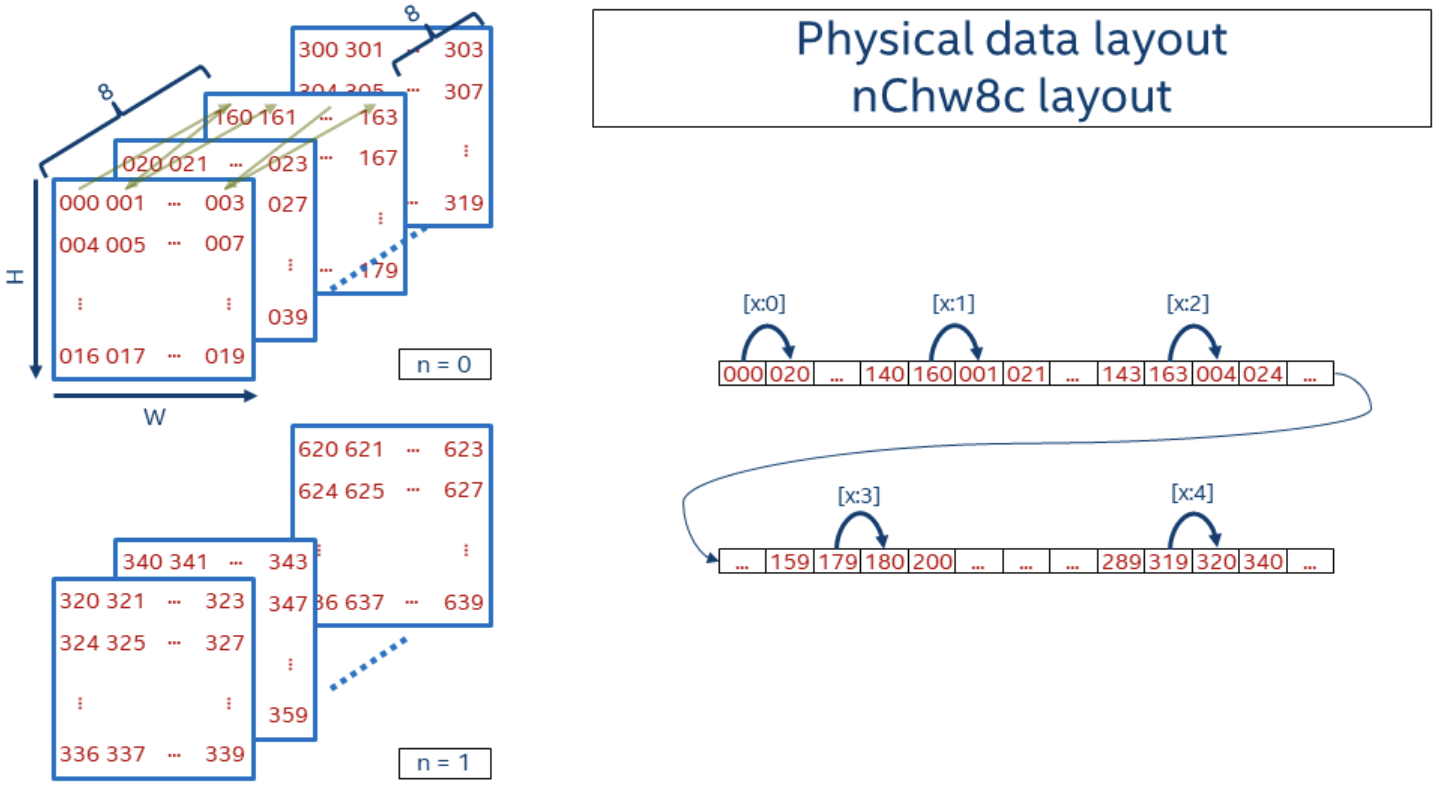}
    \caption{An illustration of nch8c spatial packing. \url{https://oneapi-src.github.io/oneDNN/dev_guide_understanding_memory_formats.html}}
    \label{fig:d}
\end{figure}

\bigskip
\noindent One way to increase the efficiency of memory accesses when parallelizing work is through spatial packing. This technique involves converting kernel arrays from a 2D NCHW format to a 4D NCHWnc packed layout. In our experiment, the converted model follows the NCHW16c format, which is widely used on AVX512+ systems. This format enables a block size of 16 and allows for 4x operations for 64 channels. If the number of channels exceeds 64, we can expect even greater improvements. Additionally, in TVM's implementation, it also enables parallelism by 4 in the H dimension, which I believe further enhances the performance.

\bigskip
\noindent For the simd schedule, TVM only supports simd int8 dot product using vmlal instructions, which enables processing 4 int8 elements in 4 int32 lanes respectively. Therefore, the maximum improvement we can achieve is 16 times, explaining why the simd int8 result is comparable to the spatially packed int8 result.

\bigskip
\noindent In NHWC spatial packing, the data is WC-packed, and it only parallelizes the H axis by a factor of 4 without additional blocking like NCHW. Consequently, this strategy yields the worst performance.

\bigskip
\noindent The quantized\_interleaved schedule is highly optimized for the NHWC layout. It utilizes the int8 4x4 matrix multiplication and accumulation operation, which involves a sequence of intrinsic instructions. This function takes two arrays, A[4][K] and B[4][K], of int8 data type and produces a 4x4 matrix equal to A*B, resulting in a fourfold optimization. In addition to the simd optimization, the quantized\_interleaved schedule also parallelizes the fused NH dimension by vectoring it to 4. This explains why NHWC with quantized\_interleaved achieves significantly higher performance.

\subsubsection{Memory-Bound Tasks}

\bigskip
\noindent Quantization brings two advantages: reduced disk or memory usage and decreased bandwidth occupation. In TVM, the performance improvement with int8 precision primarily stems from the reduction in memory bandwidth. The results are in Table \ref{tab:mem}.

\bigskip
\noindent The quantization process in TVM involves two operators, one operator reads int8 values and writes fp32 values into memory, while the other operator reads fp32 values from memory and writes int8 values. The intermediate results in memory are consistently stored as fp32 to facilitate their use in registers. \url{https://discuss.tvm.apache.org/t/tf-lite-quantized-conv2d-operator-conversion/2651/30} Besides, we always need to use fp32 to save the scale factor to preserve its precision and ensure accurate scaling during dequantization, so it is important to save the intermediate results in fp32. This also explains why memory usage remains consistent across different precisions.

\begin{table}[ht]
       \centering
        \caption{\textbf{ResNet-18 inference with batch size 64 and 256 under the best layout and schedule setup.} Here, the improvement refers to the gain achieved by using int8 precision compared to fp32 precision when all other settings remain the same. When the batch size is relatively large, the benefits of using less memory bandwidth become more apparent. \\}
        \label{tab:mem}
        \begin{tabular}{@{}ccccc@{}}
            \toprule
            \textbf{\begin{tabular}[c]{@{}c@{}}Batch Size\end{tabular}}     & \textbf{Memory (MiB)}   & \textbf{\begin{tabular}[c]{@{}c@{}}Precision\end{tabular}}    & \textbf{Time (ms)} & \textbf{Improvement}   \\
            \midrule
            \multirow{2}{*}{1}                                & 5279  & fp32 & 13.29  &   100\% \\
            & 5331 & int8 & 8.27 & 160.70\% \\

            \midrule
            \multirow{2}{*}{64}      & 5922 & fp32  & 19.65  & 100\%    \\

            & 6009   & int8 & 11.99  &   163.88\%  \\

            \midrule
            \multirow{2}{*}{256}      & 9643 & fp32  & 22.15  & 100\%    \\

            & 10061   & int8 & 11.36  &  194.98\%   \\

           \bottomrule
        \end{tabular}
        \end{table}

\end{document}